\pdfoutput=1

\documentclass[11pt]{article}

\usepackage[final]{acl}

\usepackage{times}
\usepackage{latexsym}
\usepackage{amsmath}
\usepackage{amssymb}
\usepackage{natbib}
\usepackage[T1]{fontenc}

\usepackage[utf8]{inputenc}

\usepackage{microtype}

\usepackage{inconsolata}

\usepackage{graphicx}

\usepackage{multirow}
\usepackage{booktabs}

\let\origfootnote\footnote
\renewcommand{\footnote}[1]{\kern.1em\origfootnote{#1}}
\newcommand{\punctfootnote}[1]{\kern-.1em\origfootnote{#1}}

\title{Atyaephyra at SemEval-2025 Task 4: Low-Rank Negative Preference Optimization}

\author{Jan Bronec \and Jindřich Helcl \\
Charles University, Faculty of Mathematics and Physics \\
Institute of Formal and Applied Linguistics \\
Malostranské náměstí 25, 118 00 Prague, Czech Republic \\
  \texttt{janbronec00@gmail.com}, \texttt{helcl@ufal.mff.cuni.cz}
}

\begin{document}
\maketitle
\begin{abstract}
We present a submission to the SemEval 2025 shared task on unlearning sensitive content
from LLMs. Our approach employs negative preference optimization using low-rank adaptation.
We show that we can utilize this combination to efficiently compute additional regularization terms,
which help with unlearning stabilization. 
The results of our approach significantly exceed the shared task baselines.

\end{abstract}

\section{Introduction}

Transformer-based Large Language Models (LLM) %
trained on large data corpora have shown remarkable performance in many natural language processing tasks.
However, their ability to remember portions of the training data \cite{274574} might raise legal, ethical, and other issues.
These consist of LLMs regurgitating copyright-protected creative content learned from the Web or private personal information such as social security numbers, addresses, and others.
For the latter, regulations such as the EU's General Data Protection Regulation \cite{RTBF} or the California Consumer Privacy Act \cite{CCPA} 
mandate the right for the removal of such information from the training data as per the ``Right to be forgotten''. 

Unfortunately, these issues are often discovered only after model training.
Although discarding such sensitive items from the training data sets and subsequent retraining is possible, it is generally prohibitively expensive. 
The field of machine unlearning %
tackles this exact issue by treating the removal of information as model fine-tuning.
Although several state-of-the-art approaches and benchmarks exist for LLM unlearning \cite{zhang2024negativepreferenceoptimizationcatastrophic, maini2024tofutaskfictitiousunlearning}, 
the field is still relatively unexplored.

To facilitate further progress in the field, \citet{Task} developed a comprehensive evaluation challenge
for unlearning sensitive datasets in LLM as a part of the International Workshop on Semantic Evaluation.\punctfootnote{\url{https://llmunlearningsemeval2025.github.io/}}
This paper presents our submission to the shared task.

To solve the task, we utilized the state-of-the-art unlearning method negative preference optimization
(NPO;  \citealp{zhang2024negativepreferenceoptimizationcatastrophic}).
We combined this method with low-rank adaptation (LoRA;  \citealp{hu2021loralowrankadaptationlarge})
because we consider the computational effectiveness of unlearning to be of high importance.
Although several approaches use parameter-efficient methods for unlearning in transformers \cite{gao2024practicalunlearninglargelanguage, 10677950, ding2024unifiedparameterefficientunlearningllms}, the combination of NPO and LoRA is novel.

We show that with LoRA, we can cheaply compute additional regularization terms that use the original model's output distribution without any memory overhead. 
We further show that including the KL divergence minimization regularization stabilizes the unlearning for a higher number of epochs. 
Furthermore, our solution significantly outperforms the shared task baselines.
We release the source code of our submission on GitHub.\punctfootnote{\url{https://github.com/XelfXendr/peft_unlearning}}

\begin{figure*}[ht]
    \centering
    \includegraphics[width=0.8\textwidth]{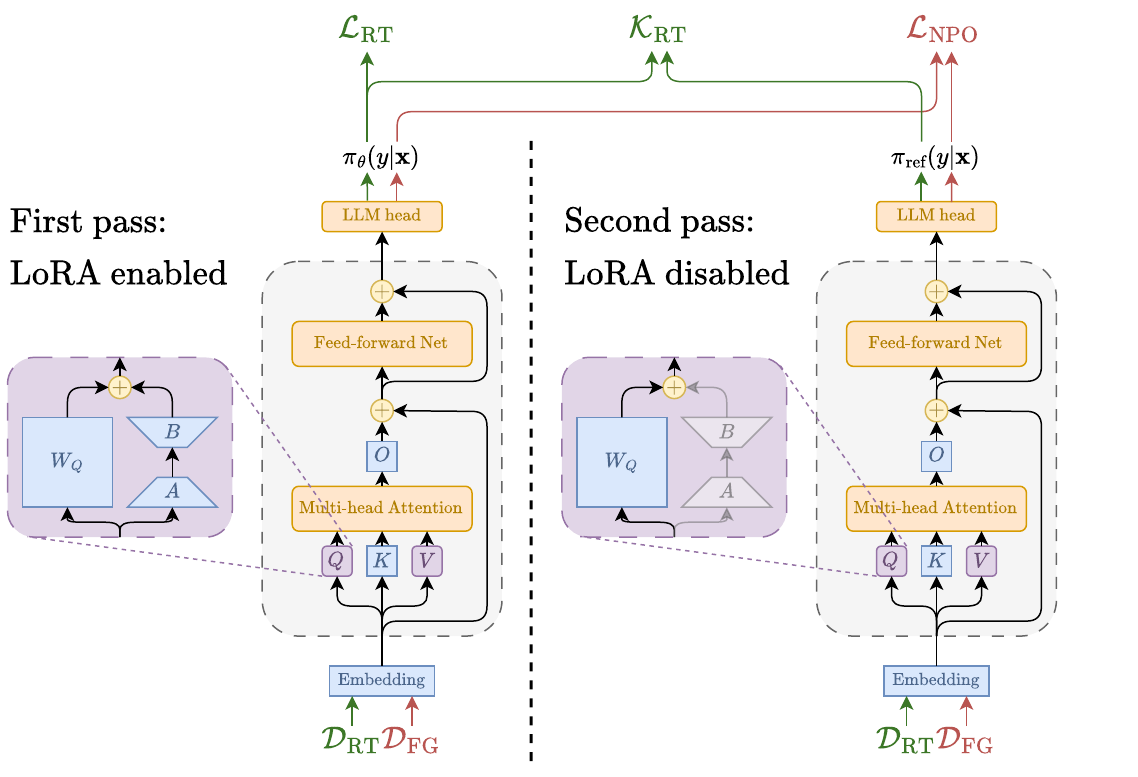}
    \caption{Overview of our unlearning method. 
    We augment the model with LoRA and train only the LoRA parameters. 
    For each batch, we compute our training loss shown in Equation \ref{eq:final_loss} in two passes.
    During the first pass, we leave the LoRA layers enabled and compute the $\mathcal{L}_\text{RT}$ term of our loss. During the second pass, we compute the $\mathcal{K}_\text{RT}$ and $\mathcal{L}_\text{NPO}$ terms by disabling the LoRA layers and only utilizing the backbone. This way, we do not need to maintain a separate copy of the backbone.
    }
    \label{fig:method_overview}
\end{figure*}
\section{Task Background}

The goal of the shared task \cite{Task} is to build a method 
for unlearning information from a given target large language model. 
For a given model, a forget set $\mathcal{D}_{\text{FG}}$, and a retain set $\mathcal{D}_{\text{RT}}$, the method should be able to remove the information present in the forget set from the model
while preserving the data from the retain set and not deteriorating the model's performance on unrelated tasks.

As the target model, the task organizers 
utilized OLMo (a pre-trained LLM), specifically its 7B and 1B versions \cite{groeneveld-etal-2024-olmo}. 
Since OLMo is trained on an open dataset Dolma \cite{soldaini-etal-2024-dolma}, it makes for a good choice for this task.

These target models were further fine-tuned to remember a dataset consisting of three document types: long-form synthetic creative documents, short-form synthetic biographies containing personally identifiable information such as fake names, phone numbers, or home addresses, and real documents sampled from the Dolma dataset. 
Each entry in the dataset consists of input-output pairs that cover either sentence completion or question answering. 

The organizers split this dataset into separate retain and forget sets, released the training and validation versions of each for the task, and provided an unlearning evaluation framework LUME \cite{ramakrishna2025lumellmunlearningmultitask}.
The data is in English only.

\section{Method Overview}

Our approach for this task combines negative preference optimization (NPO;  \citealp{zhang2024negativepreferenceoptimizationcatastrophic}) and low-rank adaptation (LoRA;  \citealp{hu2021loralowrankadaptationlarge}) for parameter-efficient fine-tuning.

\subsection{Negative Preference Optimization}

Most currently used unlearning approaches are based on gradient ascent (GA).
Consider a language model $\pi_\theta$ with parameters $\theta$, which models the next token $y$ distribution based on context $x$.
The basic premise is to ascend the classic next-token prediction cross-entropy loss on the forget data $\mathcal{D}_{\text{FG}}$ instead of descending it:
\begin{equation}
    \mathcal{L}_{\text{GA}}(\theta) = \mathbb{E}_{\mathcal{D}_{\text{FG}}}\left[ \log \pi_\theta(y|x) \right]
\end{equation}

\citet{zhang2024negativepreferenceoptimizationcatastrophic}
showed that the basic gradient ascent quickly deteriorates the utility of the model and proposed NPO
as an alternative unlearning strategy. For the original model $\pi_{\text{ref}}$, 
and a positive hyper-parameter $\beta$, the NPO loss is as follows:
\begin{multline}\label{eq:l_npo}
    \mathcal{L}_{\text{NPO}}(\theta; \beta) = \\
    = \mathbb{E}_{\mathcal{D}_{\text{FG}}}\left[ \frac{2}{\beta} \log \left( 1 + \left( \frac{\pi_\theta(y|x)}{\pi_{\text{ref}}(y|x)} \right)^\beta \right)  \right]
\end{multline}

The authors further show that $\nabla_\theta \mathcal{L}_{\text{NPO}}(\theta; \beta)$ converges to $\nabla_\theta \mathcal{L}_{\text{GA}}(\theta)$ as $\beta$ approaches zero.
For positive values of $\beta$, the NPO loss effectively dampens the contribution of already unlearned samples.

We further extend $\mathcal{L}_{\text{NPO}}$ with two regularization terms.
\citet{zhang2024negativepreferenceoptimizationcatastrophic}
regularize the NPO loss with a ``retain loss'' $\mathcal{L}_{\text{RT}}(\theta)$ to improve their unlearning results.
\begin{equation}\label{eq:l_rt}
    \mathcal{L}_{\text{RT}}(\theta) = -\mathbb{E}_{\mathcal{D}_{\text{RT}}}\left[ \log \pi_\theta(y|x) \right]
\end{equation}

As a complement to the unlearning regularization by $\mathcal{L}_{\text{RT}}$, we also utilized the Kullback-Leibler divergence $\mathcal{K}_{\text{RT}}$  between $\pi_\theta$ and $\pi_{\text{ref}}$ on the retained set. This regularization was also used by \citet{maini2024tofutaskfictitiousunlearning}. 

\begin{equation}\label{eq:k_rt}
    \mathcal{K}_{\text{RT}}(\theta) = \mathbb{E}_{\mathcal{D}_{\text{RT}}}\left[ KL\left( \pi_\theta(\cdot|x) || \pi_{\text{ref}}(\cdot|x) \right) \right]
\end{equation}
The final loss $\mathcal{L}(\theta)$ we used for the task is as follows:
\begin{multline}\label{eq:final_loss}
    \mathcal{L}(\theta; \beta, \gamma, \delta) = \\
    = \mathcal{L}_{\text{NPO}}(\theta; \beta) + \gamma\mathcal{L}_{\text{RT}}(\theta) + \delta\mathcal{K}_{\text{RT}}(\theta)
\end{multline}

A significant issue with this approach is that $\mathcal{L}_{\text{NPO}}$ and $\mathcal{K}_{\text{RT}}$
both require us to maintain the output log-probs of the original $\pi_{\text{ref}}$. 
In the case of $\mathcal{L}_{\text{NPO}}$, we only need the log-prob of the specific token in the training set, 
which we can pre-compute before unlearning.
Unfortunately, we need the entire output token distribution for $\mathcal{K}_{\text{RT}}$, 
which is unfeasible to precompute and save for each token in a sufficiently large training set.
Therefore, we must compute it on demand, for which we would typically need to keep a copy of the original model in memory.

Our contribution comes from the insight that with LoRA, 
we can circumvent the need to keep a copy of the original model
because the original model weights are still present within the augmented model.
When we need to compute $\pi_{\text{ref}}(\cdot|x)$ for $\mathcal{K}_{\text{RT}}$,  
we ignore the LoRA transformations.
We show the overall workflow of our method in Figure \ref{fig:method_overview}.
In addition to that, LoRA substantially reduces the memory requirements for fine-tuning using AdamW \cite{loshchilov2019decoupledweightdecayregularization}.

\subsection{Low-Rank Adaptation}
The individual attention layers of OLMo each contain four linear transformations
$W_q$, $W_k$, $W_v$, $W_o \in \mathbb{R}^{d\times d}$. 
The width and height $d$ of the matrices are 2048 for OLMo-1B and 4096 for OLMo-7B. 
The AdamW optimizer stores the gradient moment estimations of these matrices, which poses a significant memory cost for model fine-tuning.

The idea of LoRA is to enhance some of the linear transformations $y := W x$ 
within the attention layers of an LLM with a decomposed low-rank linear transformation.
For matrices $A \in \mathbb{R}^{r\times d}, B \in \mathbb{R}^{d \times r}$ the augmented transformation becomes $y := Wx + \frac{\alpha}{r}BAx$.
The value $r$ is the rank of the transformation, and $\alpha$ is a hyper-parameter constant in $r$.
The factor $\frac{\alpha}{r}$ is often introduced to counteract the effect that increasing $r$ has on the effective learning rate.
The original weights $W$ are then frozen during fine-tuning and only the matrices $A, B$ are updated.
This drastically decreases the number of trained parameters as $r$ can generally be set to a fairly low value, such as 2 or 5.
The memory requirements of AdamW are thus significantly reduced as well.

LoRA has a further benefit over other parameter-efficient fine-tuning methods, such as adapters %
\cite{pmlr-v97-houlsby19a} and Quantized Side Tuning \cite{zhang-etal-2024-quantized}
in that the we can merge the low-rank matrices into the original weight matrix $W' := W + \frac{\alpha}{r}BA$ after we finish fine-tuning. 
This allows us to update the model without affecting its architecture.

\newcommand{\s}{\small}

\begin{table*}[ht]
    \centering
    \begin{tabular}{cccccccccc}
        \toprule
        \multirow{2}{*}{Run} & \multirow{2}{*}{Epoch} & \multicolumn{2}{c}{Task score $\uparrow$} & \multicolumn{2}{c}{MIA score $\uparrow$} & \multicolumn{2}{c}{MMLU $\uparrow$} & \multicolumn{2}{c}{Final score $\uparrow$} \\
        \cmidrule(lr){3-4}\cmidrule(lr){5-6}\cmidrule(lr){7-8}\cmidrule(lr){9-10}
        & & $\mu$ & $\sigma$ & $\mu$ & $\sigma$ & $\mu$ & $\sigma$ & $\mu$ & $\sigma$ \\
        \midrule
        \multirow{2}{*}{$\gamma=1, \delta=0.0$} & 10 & .431 & \s .020 & .657 & \s .269 & .461 & \s .010 & \textbf{.516} & \s .089 \\
                                                & 20 & .449 & \s .029 & .389 & \s .117 & .449 & \s .008 & .429 & \s .043 \\
        \multirow{2}{*}{$\gamma=1, \delta=0.5$} & 10 & .349 & \s .045 & .375 & \s .193 & .462 & \s .005 & .395 & \s .069 \\
                                                & 20 & .434 & \s .021 & .594 & \s .201 & .439 & \s .017 & .489 & \s .074 \\
        \multirow{2}{*}{$\gamma=1, \delta=1.0$} & 10 & .349 & \s .041 & .165 & \s .083 & \textbf{.465} & \s .005 & .327 & \s .027 \\
                                                & 20 & \textbf{.453} & \s .016 & .620 & \s .219 & .449 & \s .016 & .507 & \s .072 \\
        \multirow{2}{*}{$\gamma=0, \delta=1.0$} & 10 & .369 & \s .080 & \textbf{.699} & \s .170 & .441 & \s .020 & .503 & \s .056 \\
                                                & 20 & .332 & \s .085 & .400 & \s .113 & .370 & \s .054 & .367 & \s .021 \\
        \midrule
        Baseline NPO                           & -- & .021 &  --  & .080 &  --  & .463 &  --  & .188 &  --  \\
        Baseline GD                            & -- & 0    &  --  & .382 &  --  & .348 &  --  & .243 &  --  \\
        \textit{Original model} & \textit{0} & \textit{0} & -- & \textit{0} & -- & \textit{.510} & -- & \textit{.170} & -- \\
        \bottomrule
    \end{tabular}
    \caption{Unlearning results on fine-tuned OLMo-7B-0724-Instruct-hf for various hyper-parameters. The first four entries correspond to our experimental runs with different hyper-parameter choices. Each run was repeated five times with different seeds. We report the means and standard deviations of the scores after the 10th and 20th epochs, estimated from those five runs. We compare our results to the baselines reported by task organizers. "GD" stands for Gradient Difference \cite{pmlr-v199-liu22a}. 
    }
    \label{tab:results}
\end{table*}

\begin{figure*}[ht]
    \centering
    \includegraphics[width=\textwidth]{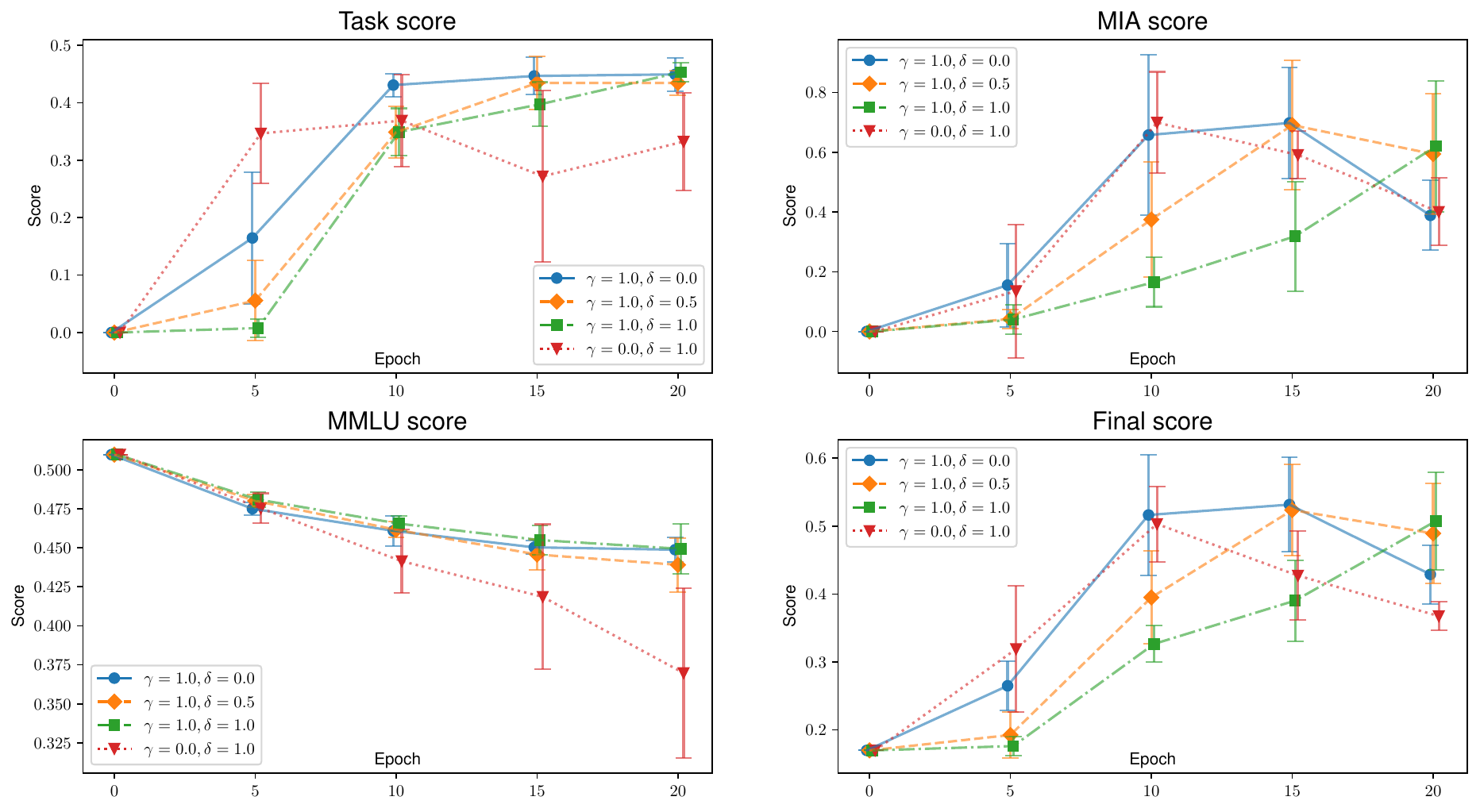}
    \caption{Unlearning results on fine-tuned OLMo-7B-0724-Instruct-hf for various hyper-parameters. Measured scores are averaged over five randomly seeded runs. The standard deviation estimates are shown in error bars. (points are offset for better visibility)}
    \label{fig:7b-scores}
\end{figure*}

\section{Experiments}\label{sec:experiments}

For our experiments, we focused on the fine-tuned OLMo-7B model, 
which we unlearned using the training retain and forget sets, 
all provided by the task organizers.
We chose a fixed value $\beta = 0.5$ of the NPO loss hyper-parameter 
and $r = \alpha = 5$ for LoRA, as these values gave us reasonably good results.
We experimented with various values for the hyper-parameters $\gamma$ and $\delta$.
The learning rate was set to $10^{-4}$ with a batch size of 4.
We perform a broader hyper-parameter search on the OLMo-1B model in Appendix \ref{sec:appendix}.

We used four combinations of values for $\gamma, \delta$. 
In three of them, we set $\delta = 1$ and experimented with values $0, 0.5,$ and $1$ for $\gamma$
to determine the effect of $\mathcal{K}_{\text{RT}}$ on the deterioration of the model. 
In the last run, we only kept $\mathcal{K}_{\text{RT}}$ with $\delta = 1$ and set $\gamma = 0$.
We ran five independent runs with unique seeds for each of the combinations.

Following the shared evaluation scheme, we measure four scores to evaluate the quality of our solution. 
First, we calculate the ROUGE-L score \cite{lin-2004-rouge} for each sample in the validation sets.
By computing the score separately for sentence-completion and question-answering pairs
of each document type in the forget and retain sets, we obtain 12 values.
We invert the values for the forget sets and produce the Task score by merging the 12 resulting values using the harmonic mean. 

We perform a loss-based Membership Inference Attack (MIA; \citealp{duan2024do}) and scale the resulting MIA score $S_{\text{MIA}}$ to penalize both under- and over-training: $S_{\text{MIA}}' := 1 - 2\cdot|S_{\text{MIA}} - 0.5|$. 
To measure the degradation of the model after unlearning, we use the MMLU benchmark \cite{hendrycks2021measuringmassivemultitasklanguage}. 
Finally, we compute the Final score as the arithmetic mean of the previous three scores.

\section{Results}

In Table \ref{tab:results}, we report the mean values and standard deviations of the different scores
estimated over the five unlearning runs. We evaluated the scores using the provided validation sets.
We show the results after 10 and 20 unlearning epochs.
We compare our results with the baselines provided by the shared task organizers.\punctfootnote{\url{https://llmunlearningsemeval2025.github.io/}} 
In Figure \ref{fig:7b-scores}, we show the development of the different scores throughout the epochs.

In general, our runs considerably outperform the provided baseline solutions. 
The runs that used $\mathcal{L}_{\text{RT}}$ overall did not degrade the MMLU score as much as those where only $\mathcal{K}_{\text{RT}}$ was involved.
Including $\mathcal{K}_{\text{RT}}$ with positive $\delta$ in the loss slowed the training. 
However, since achieving a high MIA score required hitting a sweet spot between under- and over-training,
including $\mathcal{K}_{\text{RT}}$ helped stability in the long run.

We placed lower than expected on the 7B model in the task leaderboard. 
This was caused by a logical error in our submission script, which effectively caused our solution to perform only three unlearning epochs instead of 20.
We fixed our script for the 1B model evaluation, on which we ranked as expected.

\section{Conclusion}

The combination of NPO and LoRA proved to be an effective strategy for LLM unlearning.
In addition to significant memory usage improvements, LoRA allows us to cheaply compute an additional regularization term, stabilizing the unlearning for a higher number of epochs.
In our future work, we wish to further investigate the effect of LoRA and its rank on the resilience of the model to quality deterioration.

\section*{Acknowledgments}

We would like to thank Jindřich Libovický for his comments on the draft of the paper. 

\vspace{1em}

\noindent
This work was supported by the project ``Human-centred AI for a Sustainable and Adaptive Society'' (reg. no.: CZ.02.01.01/00/23\_025/0008691), co-funded by the European Union, and by the Charles University project PRIMUS/23/SCI/023.

Computational resources were provided by the e-INFRA CZ project (ID:90254),
supported by the Ministry of Education, Youth and Sports of the Czech Republic.

\bibliography{custom}

\appendix

\section{Additional experiments}
\label{sec:appendix}

To choose the parameters for our task submission, we performed two hyper-parameter searches on the provided OLMo-1B model.
For the first search, we initially set $r = \alpha = 5$ for the LoRA parameters and searched for optimal values of parameters $\gamma, \delta$ from our training loss shown in Equation \ref{eq:final_loss}.
We experimented with utilizing only one of the regularization terms by setting one of the $\gamma, \delta$ parameters to zero, as well as combining the two terms with various factors.
We kept the NPO regularization parameter constant with the value $\beta = 0.5$.
We used the Adam optimizer with a learning rate of $10^{-4}$ and a batch size of 4 sequences. 
The fine-tuning ran for 20 epochs. We repeated each run five times with different seeds and averaged the resulting scores. We show the results in Table \ref{tab:1B_hyper}.

Overall, increasing $\gamma$ and $\delta$ led to an increase in the task score. However, too high values of $\gamma$ or $\delta$ led to a drop in the MIA score. On average, the combination of $\gamma = 1, \delta=0.5$ gave us the best final score. We used this combination for our task submission and for further experiments on the OLMo-7B model, which we discussed in Section \ref{sec:experiments}.

\begin{figure*}[ht]
    \centering
    \includegraphics[width=\textwidth]{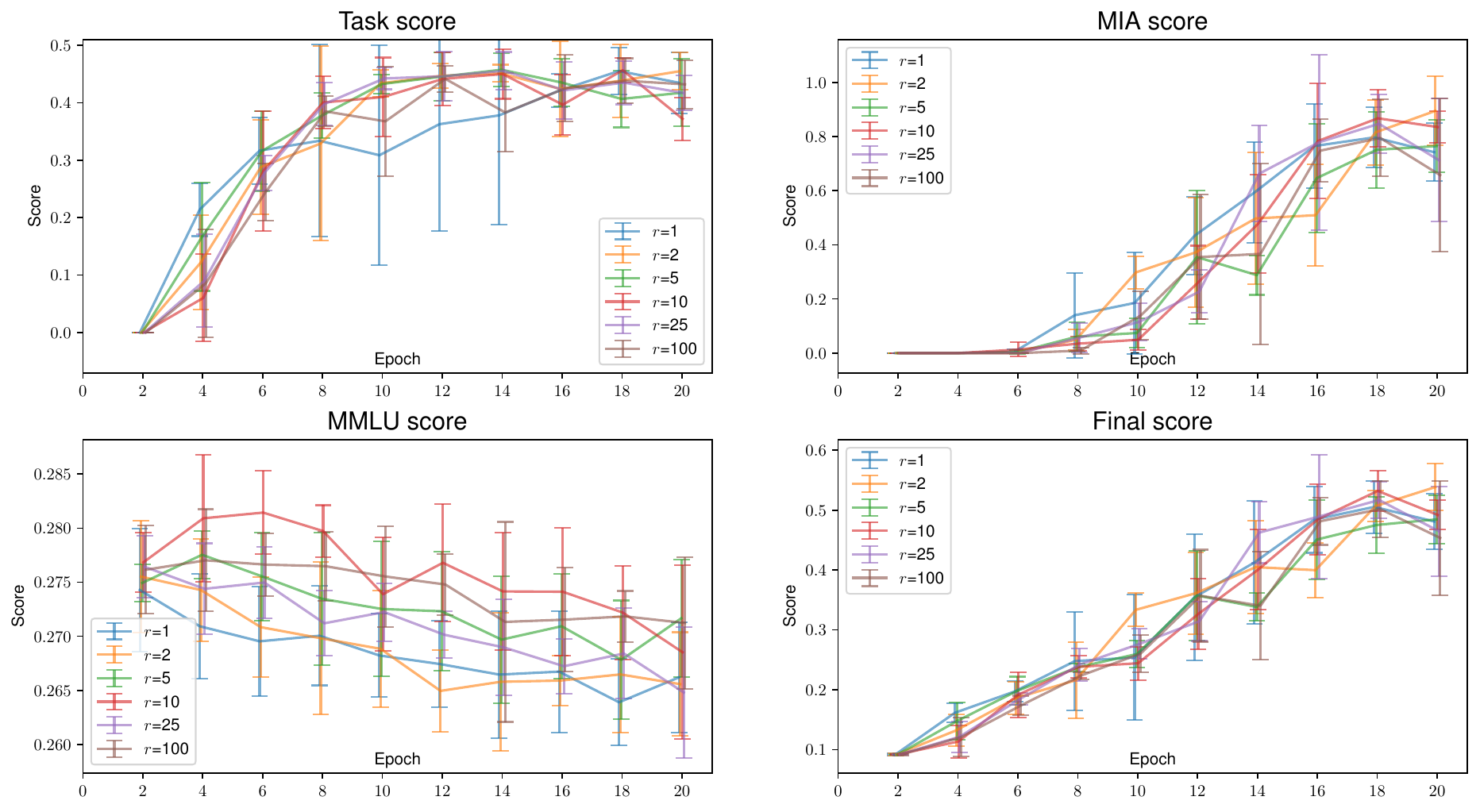}
    \caption{Unlearning results on fine-tuned OLMo-1B for various values of LoRA rank $r$. Measured scores are averaged over five randomly seeded runs. The standard deviation estimates are shown in error bars. (Points are offset for better visibility.)}
    \label{fig:rank_scores}
\end{figure*}

With the parameters $\gamma = 1, \delta=0.5$, we also performed a second hyper-parameter search for the optimal value of the LoRA rank $r$. For this search, we kept the value of $\alpha = 5$ constant, as suggested by \citet{hu2021loralowrankadaptationlarge}. We kept $\beta$, the learning rate, batch size, and the number of epochs the same as in the previous search. Once again, we ran five experiments for each value of $r$, and averaged the resulting scores. The results can be seen in Table \ref{tab:1B_rank}. In Figure \ref{fig:rank_scores}, we show the development of the scores throughout the epochs.
Although some of the runs exhibit better average performance in the various scores, we find the variance in our results too high to draw conclusions.
Ultimately, we settled with the rank $r = 5$ for our submission, as we did not see any benefits in increasing the number of parameters with higher ranks.

\begin{table*}[ht]
    \centering
    \begin{tabular}{cccccccccc}
        \toprule
        \multicolumn{2}{c}{Hyper-params} & \multicolumn{2}{c}{Task score $\uparrow$} & \multicolumn{2}{c}{MIA score $\uparrow$} & \multicolumn{2}{c}{MMLU $\uparrow$} & \multicolumn{2}{c}{Final score $\uparrow$} \\
        \cmidrule(lr){1-2}\cmidrule(lr){3-4}\cmidrule(lr){5-6}\cmidrule(lr){7-8}\cmidrule(lr){9-10}
        $\gamma$ & $\delta$ & $\mu$ & $\sigma$ & $\mu$ & $\sigma$ & $\mu$ & $\sigma$ & $\mu$ & $\sigma$ \\
        \midrule

        0.5 & 0.0 & .387 & \s .033 & .322 & \s .047 & .258 & \s .008 & .323 & \s .022 \\
        1.0 & 0.0 & .398 & \s .041 & .540 & \s .054 & .265 & \s .005 & .401 & \s .029 \\
        2.0 & 0.0 & .423 & \s .056 & .317 & \s .167 & \textbf{.270} & \s .003 & .337 & \s .049 \\
        0.0 & 0.5 & .269 & \s .086 & .285 & \s .033 & .256 & \s .006 & .270 & \s .035 \\
        0.0 & 1.0 & .245 & \s .066 & .516 & \s .074 & .262 & \s .006 & .341 & \s .040 \\
        0.0 & 2.0 & .299 & \s .059 & .599 & \s .141 & .269 & \s .004 & .389 & \s .050 \\
        0.0 & 5.0 & .362 & \s .065 & .039 & \s .049 & .267 & \s .007 & .222 & \s .021 \\
        0.2 & 0.2 & .379 & \s .030 & .295 & \s .056 & .256 & \s .011 & .310 & \s .019 \\
        0.5 & 0.5 & .350 & \s .066 & .464 & \s .104 & .268 & \s .006 & .361 & \s .048 \\
        0.5 & 1.0 & .442 & \s .030 & .719 & \s .056 & .261 & \s .009 & .474 & \s .025 \\
        1.0 & 0.5 & .394 & \s .036 & \textbf{.821} & \s .117 & .266 & \s .006 & \textbf{.494} & \s .038 \\
        1.0 & 1.0 & .419 & \s .061 & .721 & \s .198 & .270 & \s .004 & .470 & \s .084 \\
        1.0 & 2.0 & .400 & \s .081 & .606 & \s .202 & .265 & \s .003 & .424 & \s .071 \\
        2.0 & 1.0 & \textbf{.469} & \s .027 & .487 & \s .201 & .269 & \s .004 & .408 & \s .062 \\

        \bottomrule
    \end{tabular}
    \caption{
        Results of a hyper-parameter search for regularization parameters $\gamma$ and $\delta$ conducted on the fine-tuned OLMo-1B model. 
        Each run was repeated five times with different seeds, and we report the mean and standard deviation estimates of the scores after the 20th epoch.
        The scores were evaluated on validation sets provided for the shared task.
    }
    \label{tab:1B_hyper}
\end{table*}

\begin{table*}[ht]
    \centering
    \begin{tabular}{cccccccccc}
        \toprule
        \multicolumn{2}{c}{Hyper-params} & \multicolumn{2}{c}{Task score $\uparrow$} & \multicolumn{2}{c}{MIA score $\uparrow$} & \multicolumn{2}{c}{MMLU $\uparrow$} & \multicolumn{2}{c}{Final score $\uparrow$} \\
        \cmidrule(lr){1-2}\cmidrule(lr){3-4}\cmidrule(lr){5-6}\cmidrule(lr){7-8}\cmidrule(lr){9-10}
        $r$ & $\alpha$ & $\mu$ & $\sigma$ & $\mu$ & $\sigma$ & $\mu$ & $\sigma$ & $\mu$ & $\sigma$ \\
        \midrule

        1   & 5 & .435 & \s .053 & .743 & \s .107 & .266 & \s .005 & .481 & \s .047 \\
        2   & 5 & \textbf{.455} & \s .033 & \textbf{.895} & \s .127 & .266 & \s .005 & \textbf{.539} & \s .039 \\
        5   & 5 & .417 & \s .059 & .765 & \s .097 & \textbf{.272} & \s .005 & .485 & \s .041 \\
        10  & 5 & .372 & \s .037 & .835 & \s .058 & .269 & \s .008 & .492 & \s .025 \\
        25  & 5 & .417 & \s .030 & .713 & \s .227 & .265 & \s .006 & .465 & \s .075 \\
        100 & 5 & .432 & \s .042 & .658 & \s .282 & .271 & \s .006 & .454 & \s .096 \\
        
        \bottomrule
    \end{tabular}
    \caption{
        Results of a hyper-parameter search for LoRA ranks $r$ conducted on the fine-tuned OLMo-1B model.
        The regularization parameters were set to $\gamma=1, \delta=0.5$.
        Each run was repeated five times with different seeds and we report the mean and standard deviation estimates of the scores after the 20th epoch.
        The scores were evaluated on validation sets provided for the shared task.
    }
    \label{tab:1B_rank}
\end{table*}

\end{document}